# Single Class Universum-SVM

Sauptik Dhar[1]　　　　　　　Vladimir Cherkassky
LG AITeam　　　　　　　　　Dept. of ECE, UMN
Santa Clara, CA　　　　　　　Minneapolis, MN 55455
sauptik.dhar@gmail.com　　　　cherk001@umn.edu

## Abstract

This paper extends the idea of Universum learning [1, 2] to single-class learning problems. We propose Single Class Universum-SVM setting that incorporates a priori knowledge (in the form of additional data samples) into the single class estimation problem. These additional data samples or Universum belong to the same application domain as (positive) data samples from a single class (of interest), but they follow a different distribution. Proposed methodology for single class U-SVM is based on the known connection between binary classification and single class learning formulations [3]. Several empirical comparisons are presented to illustrate the utility of the proposed approach.

## 1    Introduction

The idea of Universum learning or learning through contradiction [1, 2] provides a formal mechanism for incorporating a priori knowledge about the application domain, in the form of additional (unlabeled) Universum samples. However, the implementation of Universum learning is *known only for classification setting*. It is not clear how to extend or modify this idea of learning through contradiction to other types of learning problems because the notion of 'contradiction' has been originally introduced for binary classification [1]. This paper extends the notion of Universum learning to *single-class learning* problems. For these problems, one can also expect to achieve improved generalization performance by incorporating a priori knowledge in the form of additional data samples from the same application domain. However, formalization of this idea requires significant effort. The first (conceptual) problem is that single-class model estimation (aka anomaly detection) represents unsupervised learning, where the notion of contradiction needs to be re-defined properly. This requires clear specification of the single-class learning itself.

Single class learning problems are common in many real-life applications, such as object recognition, anomaly detection, fraud detection, document classification etc. [3-6]. This problem can be formalized as follows [3].

***Problem Setting 1***: Given n samples drawn from an unknown probability distribution *P(x)*, find a "simple" region $\mathcal{S}$ of input space, such that the probability that a test point drawn from P lies outside of $\mathcal{S}$ equals some fixed value (i.e., pre-specified False_Negative error rate).

There are several known single class learning algorithms to solve this problem [3, 4, 6]. One popular approach is the single-class SVM algorithm [3], which estimates a binary function *f(x)* that takes the value +1 in a "*small*" region capturing most of the training data, and −1 elsewhere. Here the goal is to achieve a pre-specified false negative rate $FN\_rate = (1/n)\sum_{i=1}^{n} I(\mathbf{x}_i \notin \mathcal{S})$ on the training samples, which is expected to provide similar FN error rate for the future test samples drawn from the same distribution. As an illustration, consider the *handwritten digit recognition* problem, where the goal is to estimate a single-class decision rule for the images of the handwritten digit "0" in a 28x28 pixel space. The typical approach adopted is to estimate a single-class SVM model using the training samples of digit "0", which achieves a pre-specified user-defined *FN_rate*.

---

[1] This work was done during Dr. Sauptik Dhar's PhD at UMN.

In many applications, single class SVM is used under a different problem setting, where in addition to test samples from distribution *P,* the *test* data also contains additional samples from a *different* distribution *Q*. Typically, samples from *P* are labeled as *normal (or positive) class* and those from *Q* are labeled as *abnormal class (aka negative)*. Under this setting, the goal is to estimate a single class model which minimizes false positive rate *FP_rate* for a given *FN_rate*. Note that negative samples are not available (known) during training. For the digit recognition example, the normal class may represent samples of digit "0" and the abnormal class may constitute samples from other digits "1-9". This approach is implicitly adopted in many applications [4, 5, 7, 8, 9]. Unfortunately, this approach is fundamentally flawed, in that *only positive samples* are available (known) during training a single-class model. So it is not possible, in principle, to achieve the goal of minimizing the FP error rate for test samples from a completely unknown distribution *Q*. This situation breeds many heuristic algorithms for single-class learning that exhibit 'superior' performance simply because an algorithm is a better match for specific data sets [4, 7, 8, 9].

This paper introduces a better setting for single class learning, where, in addition to positive training samples, one has 'Universum' samples from distribution *U* which is different from *P*. Universum samples belong to the same application domain and are defined in the same input space (x) as training and test samples. Therefore, Universum data *may* provide useful information about unknown *P* and *Q*, and so it can help minimize the number of FP errors for test samples from *Q*. For example, in the handwritten digit recognition problem, training data may include positive training samples of digit "0", Universum samples of handwritten letters (in the same 28x28 pixel space), and test samples of other digits "1-9" (~ negative class). The goal of learning is to estimate a single class model which minimizes FP_rate for a given FN_rate. Under this setting, a single-class model is estimated from positive (normal) training samples *and* Universum samples. Even though Universum samples (letters) do not belong to the normal class, they reflect the style of handwriting, and can be possibly useful for minimizing FP error rate for 'other digits' from unknown distribution *Q*. This can be formalized as the following setting for single-class learning.

***Problem Setting 2****: Given normal (positive) samples drawn from unknown distribution P(**x**), and Universum samples from unknown distribution U(**x**), find a "simple" region $\mathcal{S}$ of input space, that provides pre-specified FN error rate and minimum FP error rate for future test samples. It is assumed that test samples are generated from some mixture of P(**x**) and Q(**x**), aka positive (normal) and negative (abnormal) class distributions.*

Under this setting, additional samples from *U(**x**)* contain additional a priori knowledge about application domain, similar to Universum learning originally introduced for binary classification setting [1, 2]. However, the idea of 'learning through contradiction' [1, 2] cannot be easily extended to single-class learning, where negative samples are not available for training (model estimation). This paper shows how to incorporate Universum samples into SVM-like formulation for single class learning, in order to solve *Problem Setting 2*. This new Single Class U-SVM formulation builds upon standard single-class SVM formulation [3] providing solution to *Problem Setting 1*.

The paper is organized as follows. Section 2 reviews the connection between single-class SVM and standard SVM classification developed in [3]. This connection is later used for introducing Universum into single class learning. Section 3 describes the proposed single-class U-SVM formulation, as a constructive solution for single-class learning (under Setting 2). Section 4 presents empirical results for single-class U-SVM. Finally, conclusions are presented in Section 5.

## 2     Single Class SVM via Binary Classification

Single-class SVM formulation [3] was proposed for solving Problem Setting 1. For improved readability, we show only linear SVM parameterization (see Algorithm 1). It can be readily extended to nonlinear SVM via kernels [3, 6].

**Algorithm1**: Single class SVM optimization

A1. Given, training data from the *"normal class"* $\{\mathbf{x}_i\}_{i=1}^{n}$

A2. Solve the following optimization problem,

$$\min_{\mathbf{w},\rho,\xi} \quad \frac{1}{2}\|\mathbf{w}\|^2 + \frac{1}{\nu n}\sum_{i=1}^{n}\xi_i - \rho \quad (1)$$

s.t. $\quad (\mathbf{w}\cdot\mathbf{x}_i) \geq \rho - \xi_i, \quad \xi_i \geq 0, \quad i = 1 \text{ to } n$

A3. Finally, predict on future test data as $\quad D(\mathbf{x}_i,\mathbf{w},\rho) = \begin{cases} +1\ (\text{"normal"}) & \text{if } (\mathbf{w}\cdot\mathbf{x}_i) \geq \rho \\ -1\ (\text{"abnormal"}) & \text{otherwise} \end{cases}$

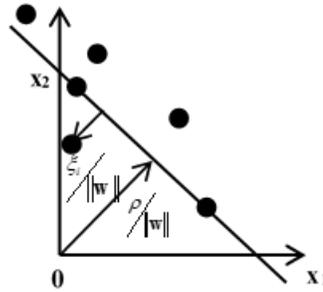

Figure 1: Schematic representation of single-class SVM optimization.

Under this approach, a large margin ($\rho$) hyperplane, characterized by a margin size $\rho/\|\mathbf{w}\|$, is used to separate the training samples from the origin (as shown in Fig.1). Samples lying outside this margin are linearly penalized using the slack variables $\xi_i$. The tradeoff between margin size and margin errors is controlled by the user defined parameter $\nu \in [0,1]$. This parameter $\nu$ also acts as the lower bound on the fraction of training points that are support vectors and controls the FN_rate for training samples. This indirectly controls the FN_rate for test samples (from the same distribution). Typically, a small value of $\nu$ is likely to provide a small FN_rate on the training as well as the future test samples. Selecting this parameter is application dependent and is set (by a user) based on application domain knowledge [6].

Schölkopf and Smola [3], proposed to map a single-class SVM problem onto an 'equivalent' binary SVM classification problem, as shown in Fig. 2(a), (b). Their proposition 8.2 shows how the margin of an equivalent binary SVM classifier is related to the single-class SVM decision boundary:

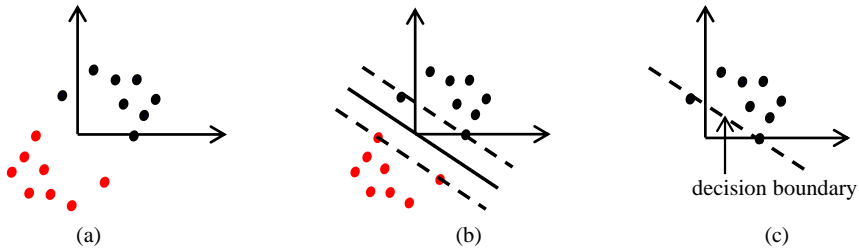

Figure 2: A schematic representation of solving single-class SVM problem using a binary SVM classifier. (a) Reflect training samples about the origin. (b) Estimate a binary SVM classifier. (c) Predict on future test data using the margin as the single-class decision boundary.

**Proposition 8.2 [3]**: *Suppose* $(\mathbf{w}, b = 0)$ *parameterizes the optimal separating hyperplane passing through the origin for a labeled data set* $\{(\mathbf{x}_i, y_i = \pm 1)\}_{i=1}^{n}$ *aligned such that* $(\mathbf{w}\cdot\mathbf{x}_i)$ *is positive for* $y_i = +1$. Suppose, moreover, that $\rho/\|\mathbf{w}\|$ is the margin of the optimal hyperplane. Then $(\mathbf{w}, \rho)$ parameterizes the supporting hyperplane for the unlabeled data set $\{y_i \mathbf{x}_i\}_{i=1}^{n}$.

**Proof**: See [3].

This connection yields the following Algorithm 2 for solving single-class SVM via binary SVM classification. (A schematic representation of this algorithm is shown in Fig. 2).

---
**Algorithm 2**: Solving single class SVM problem using a binary SVM classifier (see Fig. 2)

B1. Given, training data from the *"normal class"* $\{\mathbf{x}_i\}_{i=1}^{n}$. Reflect the data about the origin and label them as shown (see Fig. 2a), $\mathbf{x}_i = \mathbf{x}_i$ with $y_i = +1$, for $i = 1$ to $n$

$\qquad = -\mathbf{x}_i$ with $y_i = -1$, for $i = n+1$ to $2n$

B2. Solve the binary SVM problem

(2)
$$\min_{\mathbf{w},b,\xi} \quad \frac{1}{2}\|\mathbf{w}\|^2 + C\sum_{i=1}^{2n}\xi_i$$

$$\text{s.t} \quad y_i(\mathbf{w}\cdot\mathbf{x}_i + b) \geq 1-\xi_i \quad \xi_i \geq 0 \quad i = 1 \text{ to } n$$

B3. The single class SVM decision rule is given as:
$$D(\mathbf{x}_j, \mathbf{w}) = \begin{cases} +1 \text{ ("normal") ; if } (\mathbf{w}\cdot\mathbf{x}_j) \geq 1 \\ -1 \text{ ("abnormal") ; otherwise} \end{cases}$$

---

Here the parameter C is equivalent to the parameter $\nu$, and controls the *FN_rate* of the training as well as the future test samples. Note that, for improved readability, we show only linear parameterization in Algorithm 2. However (2) can be generalized to the nonlinear case by solving the problem in the dual form and using the kernel $\mathbf{K} = \begin{bmatrix} \mathbf{K}_T & -\mathbf{K}_T \\ -\mathbf{K}_T & \mathbf{K}_T \end{bmatrix}$, where $\mathbf{K}_T$ is the kernel for the training samples. This modified kernel $\mathbf{K}$ captures the effect of reflecting the training samples about the origin in the kernel space.

Note that Algorithm 2 effectively shows how to solve single-class Setting 1 via binary SVM classification. This connection enables application of existing SVM classification software for single-class Setting 1. In the next section, we use this connection to introduce the new single-class Universum SVM formulation.

## 3  Single Class U-SVM

Next, we introduce our new formulation called the single-class Universum support vector machine (U-SVM) used to solve Problem Setting 2. Under this setting we are given training samples from normal class and a set of examples from the Universum. The Universum contains data that belongs to the same application domain as the training data, but these samples are known not to follow the same distribution as the "normal" class. Note however that the Universum samples *may or may not* follow the same distribution as the abnormal class. These Universum samples are incorporated into single-class learning as explained next. Let us assume that the training samples are linearly separable using two large margin hyperplanes $\mathcal{H}_1$ and $\mathcal{H}_2$ with the same margin size. Then the Universum samples can fall on either side of the decision boundary (see Fig. 3). Note that, we should favor hyperplane models where the Universum fall on the wrong side of the decision boundary (i.e., they should not be classified as the normal class). Such Universum samples are called contradictions, because they are falsified by the model (i.e., have nonzero slack variables). Thus, Universum learning implements a trade-off between explaining training samples from the normal class (using large margin hyperplanes) and maximizing the number of contradictions (on the Universum).

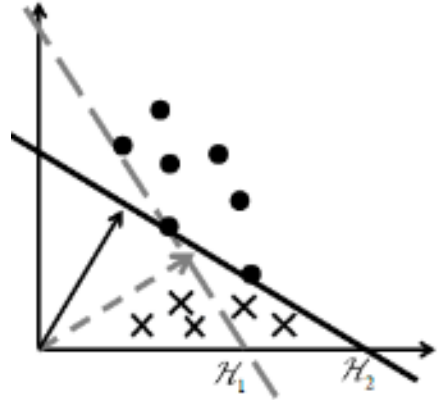

Figure 3: Two large margin hyperplane with different number of contradictions on the Universum. Bold Dots are training samples and crosses are Universum samples.

The quadratic optimization formulation for implementing an SVM-style inference through contradictions is introduced next. For training samples, we use the standard single-class SVM soft-margin loss with slack variables $\xi_i$ (as in Algorithm 2). The universum samples $(\mathbf{x}_j^*)$ are penalized using a hinge loss on the universum samples $L_\varepsilon(f(\mathbf{x},\mathbf{w})) = \max((\mathbf{w}\cdot\mathbf{x}) - \varepsilon, 0)$, with $\varepsilon > 0$ (as shown in Fig. 4). This loss function forces the Universum to lie far away from the decision boundary. That is, the universum samples with $(\mathbf{w}\cdot\mathbf{x}) \geq \varepsilon$ are linearly penalized using the slack variables $\xi_j^*$. The proposed single-class U-SVM formulation is shown next:

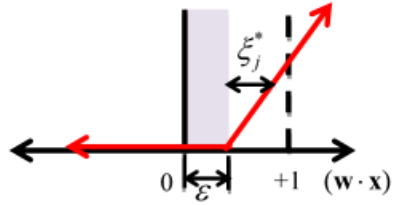

Figure 4: The hinge loss for the Universum samples.

Single-class U-SVM formulation

$$\min_{\mathbf{w},\boldsymbol{\xi}} \quad \boxed{\frac{1}{2}\|\mathbf{w}\|^2 + \hat{C}\sum_{i=1}^{n}\xi_i} \quad + \quad \boxed{C^*\sum_{j=1}^{m}\xi_j^*} \quad (3)$$

$$\text{s.t.} \quad \boxed{\begin{array}{l}\mathbf{w}\cdot\mathbf{x}_i \geq 1-\xi_i \\ \xi_i \geq 0, i=1 \text{ to } n\end{array}} \quad \boxed{\begin{array}{l}\mathbf{w}\cdot\mathbf{x}_j^* \leq \varepsilon + \xi_j^* \\ \xi_j^* \geq 0, j=1 \text{ to m}\end{array}}$$

(samples from "*normal*" class)   (samples from "*universum*")

Here, the user-defined parameters $\hat{C}, C^* \geq 0$ control the tradeoff between minimizing the FN_rate on training samples and maximizing the number of contradictions on the universum samples. Formulation (3) can be solved via an equivalent binary classification formulation, effectively following the same steps as in Algorithm 2 (shown in Fig.2). This leads to the following Algorithm 3 for solving (3).

The solution to the optimization problem (4) defines the large margin hyper-plane $f(\mathbf{x}) = (\mathbf{w}\cdot\mathbf{x}) - 1 = 0$ that incorporates a priori knowledge (i.e., Universum samples) into the final single-class model. Following the same arguments as in proposition 8.2[3], we have b=0. Hence (4) is equivalent to solving (3) with $\hat{C} = 2C$, except for an additional constraint $-\mathbf{w}\cdot\mathbf{x}_j^* \leq \varepsilon + \xi_j^*$ on the universum samples. Note that for Universum samples, constraint $-\mathbf{w}\cdot\mathbf{x}_j^* \leq \varepsilon + \xi_j^*$ acts as an inactive constraint and does not affect the solution. The computational cost for solving this single-class U-SVM problem is the same as solving the standard binary U-SVM with *2n* training samples and *m* universum samples. This is equivalent to solving the standard binary SVM problem with 2(n+m) samples [10, 11].

---

**Algorithm 3**: single class U-SVM Formulation.

C1. Given, training data from the *"normal class"* $\{\mathbf{x}_i\}_{i=1}^{n}$ and additional universum samples $\{\mathbf{x}_j^*\}_{i=1}^{m}$. Reflect the training data and label them as in (B1). Note, we do not reflect the universum samples.

C2. Solve the binary U-SVM problem [10, 11],

$$\min_{\mathbf{w},b,\xi} \quad \boxed{\frac{1}{2}\|\mathbf{w}\|^2 + C\sum_{i=1}^{2n}\xi_i} \quad + \quad \boxed{C^*\sum_{j=1}^{m}\xi_j^*} \quad (4)$$

$$\text{s.t.} \quad y_i(\mathbf{w}\cdot\mathbf{x}_i+b) \geq 1-\xi_i \quad \left|(\mathbf{w}\cdot\mathbf{x}_j^*)+b\right| \leq \varepsilon + \xi_j^*$$

$$\xi_i \geq 0 \ i=1 \text{ to } 2n \quad \xi_j^* \geq 0, j=1,...,m$$

(samples from "*normal*" class)     (samples from "*universum*")

C3. Finally, predict on the future test data as $D(\mathbf{x}_i,\mathbf{w}) = \begin{cases} +1 \ (\text{"normal"}) & \text{if } (\mathbf{w}\cdot\mathbf{x}_i) \geq 1 \\ -1 \ (\text{"abnormal"}) & \text{otherwise} \end{cases}$

---

The kernelized version of U-SVM formulation (4) has four tunable parameters: C, C*, kernel parameter and $\varepsilon$. So model selection (parameter tuning) becomes an issue for any real-life application. We propose the following model selection strategy for estimating a single-class U-SVM model which minimizes FP_rate for a pre-specified FN_rate (e.g., Problem Setting 3). This setting leads to the following two-step model selection strategy,
a   Fix C and kernel parameter values as appropriate for the single class SVM model (2). This provides a fixed FN_rate on the training samples and the future test samples from the "normal" class.
b   Perform model selection for the ratio C*/C in (4) while keeping C and kernel parameters fixed (as in (a)). This is achieved by choosing the maximum value of C*/C providing fixed (pre-specified) FN error rate on an independent validation set (of positive samples). Parameter $\varepsilon$ is usually pre-set to a small value and does not require tuning. For this paper we set $\varepsilon = 0$.

Further, the performance of the single-class U-SVM may also depend on the number of universum samples used. For simplicity, in this paper we keep it equal to the number of training samples. Next, we provide empirical results to show the effectiveness of the proposed single-class U-SVM over single-class SVM under Problem Setting 2.

## 4     Experimental Results

### 4.1     MNIST

For our first set of experiments we use the real-life handwritten digits recognition MNIST data [12]. The goal is to build a single-class classifier for digit "0", where each sample is represented as a real-valued vector of size 28x28=784. For illustration we show the results for handwritten digits "1" and "2" as universum. We test our estimated single-class models under two distinct scenarios. In the first case, the "abnormal" class constitutes of the handwritten digits "3 to 9". Here, the samples of the abnormal class follow a different distribution than the normal class as well as the universum samples. For the second case, the "abnormal" class contains handwritten digits "1" or "2". In this case, the abnormal class follows the same distribution as the universum. The experimental setting used for this example follows next,
- No. of training/validation samples = 1000. (digit "0"). (The validation set contains independent samples of "0", used to select the largest C*/C (ratio) which provides the same FN_rate on the validation set, as the single-class SVM).
- No. of additional Universum samples = 1000. (digit "1" and "2").
- No. of test samples: (we use all the samples available in the separate test set [12])
  - *normal* class (from digit "0") with 980 samples.
  - *abnormal* class (*for case 1:* 6853 samples from digits 3 to 9).
  - *abnormal* class (*for case 2*: 1135 samples from digit "1" ; 1032 samples from digit "2").

All experiments use linear SVM parameterization which is appropriate for this (sparse) data set. We provide our results for two different model parameters $v = 2^{-8}$, $2^{-4}$; characterizing high and low FN_rates on the training / test samples from the "normal" class. Table 1 shows performance comparison for the single-class SVM vs. single-class U-SVM. The table shows the average value of the FN_rate and FP_rate (in percent) for 10 random selection of the training/validation set. The test set is kept fixed. The standard deviation of the FN_rate and FP_rate are shown in parenthesis.

Comparisons shown in Table 1 indicate that both standard single-class SVM and the proposed U-SVM yield the same FN test error rate (as expected), but U-SVM provides lower (or, in some cases, similar) FP test error rates. Specific comparison results depend on both the value of FN error rate (pre-set by a user) *and* the quality of Universum. For cases with high values of $v = 2^{-4}$ (characterizing high FN_rate), the training data is not well-separable. For such cases, the original motivation for Universum learning, to stabilize selection of a large margin hyperplane does not work (see Fig. 3). Hence, single-class U-SVM is not likely to provide any improvement. However, for cases with high-separability of the training samples (i.e., for $v = 2^{-8}$ with low FN_rate), the single class U-SVM provides a significant improvement using digit "1". Such an improvement is not seen with digit "2" as universum. This can be (intuitively) explained by noting visual similarity between digits "2" and "0". Hence, digit "2" is not a good contradiction for digit "0".

Table 1: Comparison of single-class SVM vs. U-SVM on MNIST (digit "0" as "normal class").

|  |  | Single-SVM | Single U-SVM (digit "1") | Single U-SVM (digit "2") |
|---|---|---|---|---|
| $v = 2^{-8}$ ( equivalent C=$2^{-7}$ )   ~ Low FN rate ||||||
| Training | FN (%) | 0.7 (0.2) | 0.8 (0.2) | 0.8 (0.4) |
| Test | FN (%) | 1.8 (0.7) | 1.8 (0.6) | 1.8 (0.75) |
|  | FP (%) (digits " 3-9") | **69 (3.9)** | **56 (5.2)** | **69 (3.8)** |
|  | FP (%) (on digit " 1") | **4.1 (0.4)** | **0.6 (0.6)** | - |
|  | FP (%) (on digit "2") | **79 (3.1)** | - | **72 (2.7)** |
| $v = 2^{-4}$ (equivalent C=$2^{-11.5}$ ) ~ High FN rate ||||||
| Training | FN (%) | 5.9 (0.2) | 6.3 (1) | 6 (0.3) |
| Test | FN (%) | 6.5 (0.6) | 6.5 (0.6) | 6.5 (0.6) |
|  | FP (%) (digits " 3-9") | 43 (2.4) | 43 (3.8) | 43 (2.4) |
|  | FP (%) (on digit " 1") | 1.5 (0.5) | 1.0 (0.7) | - |
|  | FP (%) (on digit "2") | 54 (3.3) | - | 54 (3.1) |

### 4.2 Reuters 21578 v1.0

Our next set of results uses the real-life Reuters-21578 v1.0 data [13]. It consists of 21,578 news stories that appeared in the Reuters newswire in 1987, which are classified according to 135 thematic categories mostly concerning business and economy. Here we use the subset R90 of this collection and the standard *ModApte'* split to define the documents used as training and testing examples (see [13] for details). We use the preprocessed term-frequency encoded data using 5180 selected words already available in [13]. The goal is to build a single-class classifier for the category "crude". We show the results for two types of Universa: "money-fx" and "trade". As before, we consider two extreme cases for the "*abnormal class*". For the first case, the abnormal class consists of all samples from the 90 other categories; except the categories "crude" (normal class), "money-fx" or "trade" (universum). For the second case, the "abnormal" class consists only of the unseen samples from the universum (i.e. "money-fx" or "trade"). The experimental setting is detailed below,
- No. of training/ validation samples = 195. (samples from category "crude").
- No. of additional Universum samples = 200. (samples from "money-fx", "trade").
- No. of test samples: (we use all the samples available in the separate test set [13])
    - *normal* class (from category "crude") with 189 samples,
    - *abnormal* class (case 1: 2539 samples from 90 *other* categories except "crude", money-fx", "trade").
    - *abnormal* class (case 2: 517 samples from "money-fx" ; 286 samples from "trade").

All experiments use linear parameterization appropriate for this dataset. Table 2 shows empirical results, i.e. training/test error rates averaged over 10 random experiments. As shown in Table 2, for the low FN_rate ($\nu = 2^{-4}$) the U-SVM with "money-fx" provides good improvement over the single-class SVM. Here the universum "trade" does not provide significant improvement over single-class SVM.

Table 2. single-class SVM vs. U-SVM on Reuters-21578 (category "crude" as "normal class").

| | | Single-SVM | Single U-SVM ( "money-fx") | Single U-SVM ( "trade") |
|---|---|---|---|---|
| $\nu = 2^{-4}$ ( equivalent C=$2^{-7}$ ) ~ Low FN rate | | | | |
| **Training** | FN (%) | 11 (3.0) | 13 (3.4) | 11 (2.5) |
| **Test** | FN (%) | 28 (0) | 29 (2.5) | 29 (1.2) |
| | FP (%) (on *others*) | **14 (1.6)** | **8 (1.1)** | **12 (1.5)** |
| | FP (%) (on "money-fx") | **16 (4.0)** | **0.8 (0.1)** | - |
| | FP (%) (on "trade") | **29 (1.2)** | - | **15 (0.7)** |
| $\nu = 2^{-2}$ ( equivalent C=$2^{-9.5}$ ) ~ High FN rate | | | | |
| **Training** | FN (%) | 25 (1.7) | 26 (1.5) | 26 (2.8) |
| **Test** | FN (%) | 37 (2.4) | 37 (2.4) | 37 (2.2) |
| | FP (%) (on *others*) | 12 (2.2) | 12 (2.0) | 12 (2.0) |
| | FP (%) (on "money-fx") | 16 (1.0) | 15 (1.9) | - |
| | FP (%) (on "trade") | 35 (3.1) | - | 34 (3.2) |

## 5    Conclusion

This paper introduced single-class U-SVM formulation for *Problem Setting 2*. Setting 2 for single-class learning is (implicitly) adopted in many applications dealing with single-class learning and anomaly detection. The proposed single-class U-SVM can be implemented via minor modification of the binary U-SVM software [10]. Further, we provide a sound practical strategy for tuning model parameters in the proposed single-class U-SVM. Finally, we provide empirical results to show the effectiveness of the proposed single-class U-SVM over standard single-class SVM.

For most applications, there is still a need to provide better characterization of the 'good' Universa, for which U-SVM can provide improvement over the single-class SVM. Following the method of "histogram of projections" [14], similar practical conditions for the effectiveness of single-class U-SVM can be obtained. These conditions (not shown here due to space constraints) will be presented in the future.

## References

[1]    V. N. Vapnik, *Statistical Learning Theory*. New York: Wiley, 1998

[2]     V. N. Vapnik, *Estimation of Dependencies Based on Empirical Data. Empirical Inference Science: Afterword of 2006*. New York: Springer, 2006.

[3]     B. Schölkopf, and A. Smola, *Learning with Kernels*. MIT Press, 2002.

[4]     V. Chandola, A. Banerjee, and V. Kumar. "Anomaly detection: A survey," *ACM Comput. Surv.*, vol. 41, no. 3, pp. 1-58, 2009.

[5]     L. M. Manevitz and M. Yousef. "One-class SVMs for document classification," *Journal of Machine Learning Research*, vol. 2, pp. 139-154, 2002.

[6]     V. Cherkassky, and F. Mulier, *Learning from Data Concepts: Theory and Methods*, 2nd ed. NY: Wiley, 2007.

[7]     E. Eskin, A. Arnold, M. Prerau, L. Portnoy, and S. Stolfo,"A geometric frame-work for unsupervised anomaly detection". *In Proceedings of Applications of Data Mining in Computer Security*, 2002.

[8]     E. Eskin, W. Lee, and S. Stolfo, "*Modeling system call for intrusion detection using dynamic window sizes,*" In Proceedings of DISCEX, 2001.

[9]     K.A. Heller, K.M. Svore, A. Keromytis, S.J. Stolfo, "*One class support vector machines for detecting anomalous windows registry accesses,*" in: Proc. The workshop on Data Mining for Computer Security, 2003, pp. 281–289.

[10]    UniverSVM. [WWW page].URL: http://mloss.org/software/view/19/.

[11]    J. Weston, R. Collobert, F. Sinz, L. Bottou, and V. Vapnik, "*Inference with the Universum*," Proc. ICML, 2006, pp. 1009–1016.

[12]    S. Roweis, sam roweis: data. [WWW page]. URL http://www.cs.nyu.edu/~roweis/data.html

[13]    R. Correa, T. Ludermir, "A quickly trainable hybrid SOM-based document organization system," *Neurocomputing*, vol .71, pp. 3353-3359, 2008. doi:10.1016/j.neucom.2008.02.021.

[14]    V. Cherkassky, S. Dhar, and W. Dai, "Practical Conditions for Effectiveness of the Universum Learning," *IEEE Transactions on Neural Networks*, vol.22, no. 8, pp. 1241-1255, Aug 2011.